\documentclass[conference]{IEEEtran}
\usepackage{cite}
\usepackage{amsmath,amssymb,amsfonts}
\usepackage{algorithmic}
\usepackage{graphicx}
\graphicspath{{figs/}}
\usepackage{textcomp}
\usepackage{xcolor}

\newcommand{\edit}[1]{{#1}}

\def\BibTeX{{\rm B\kern-.05em{\sc i\kern-.025em b}\kern-.08em
    T\kern-.1667em\lower.7ex\hbox{E}\kern-.125emX}}
    
\begin{document}

\title{Predicting Hateful Discussions on Reddit using Graph Transformer Networks and Communal Context
}

\author{\IEEEauthorblockN{Liam Hebert}
\IEEEauthorblockA{
\textit{University of Waterloo}\\
Waterloo, Canada \\
liam.hebert@uwaterloo.ca}
\and
\IEEEauthorblockN{Lukasz Golab}
\IEEEauthorblockA{
\textit{University of Waterloo}\\
Waterloo, Canada \\
lgolab@uwaterloo.ca}
\and
\IEEEauthorblockN{Robin Cohen}
\IEEEauthorblockA{
\textit{University of Waterloo}\\
Waterloo, Canada \\
rcohen@uwaterloo.ca}
}

\maketitle

\begin{abstract}
We propose a system to predict harmful discussions on social media platforms. Our solution uses contextual deep language models and proposes the novel idea of integrating state-of-the-art Graph Transformer Networks to analyze all conversations that follow an initial post. This framework also supports adapting to future comments as the conversation unfolds. In addition, we study whether a community-specific analysis of hate speech leads to more effective detection of hateful discussions. We evaluate our approach on 333,487 Reddit discussions from various communities. We find that community-specific modeling improves performance two-fold and that models which capture wider-discussion context improve accuracy by 28\% (35\% for the most hateful content) compared to limited context models. 
\end{abstract}


\section{Introduction}
The rise of social media has led to increased harmful discourse with damaging mental health effects. A recent study by Gao et al. measured the mental health of over 4800 Chinese citizens and found a high correlation between increased anxiety and depression with high social media usage \cite{2020jungling}. These alarming trends have motivated the usage of automated detection methods to help moderate these platforms and prevent further harms.

A common approach of methods that tackle this problem is to detect the hatefulness of individual comments that belong to a wider discussion \cite{Obadinma2021Class, huang-etal-2018-cyberbullying, Brassard2021Using}. Two central challenges arise, however. First, conversational hate speech is deeply contextual. To properly understand if a comment is hateful, it is important to understand the context in which it was said. 
Second, by only analyzing the hatefulness of individual comments, we are bound to a reactive kind of moderation, which requires the damaging comment and the discussion that led to it be published first. This can lead to the spread of conversations that discuss harmful or sensitive subjects which can incite further hateful discourse. As such, it is important to proactively detect the tone and trend of conversations before they can delve into explicit hate speech.

Towards the goal of proactive detection, we also believe that it is important to consider the unique latent communal norms and behaviours that underpin social groups. Content which may trigger a hateful reaction from members of one group can trigger a different reaction from another group. This focus is inspired by a recent study by Cinellie et al., which found that online communities are prone to the ``echo chamber’’ effect, where extreme ideals can be encouraged and amplified 
\cite{Cinellie2023301118}. As such, learning these cultural norms and behaviours is essential to properly model the direction of a conversation and whether that direction leads to hate speech. 

Taking all these observations into consideration, the question is how to design a framework that can perform a rich analysis of branching social media discussions with a view of proactive hate speech prediction, and also how best to design experiments to demonstrate whether community-specific analysis provides important benefits.

To study the behaviours of diverse communities, we focus on the social platform Reddit, where discussions take place in topic-orientated communities called subreddits. These communities are highly varied, ranging from \texttt{aww}, to discuss cute animals, and \texttt{politics}, to discuss political affairs.
Beyond their subject differences, prior work by Waller et al. found that these communities exhibit significant differences in their social makeup and discussion behaviours \cite{Waller2021}. For example, discussions that take place on \texttt{The\_Donald} were found to have a significant political right leaning bias whereas \texttt{politics} has an opposing left leaning bias. Within these communities, conversations take place in branching trees structures where any user can reply to the comments of other users, forming a linked chain of replies that can branch off into sub-discussions. In addition to textual replies, users can explicitly provide their opinion by up-voting (approving) or down-voting (disapproving) the comments of other users. 

\begin{figure}
    \centering
    \includegraphics[width=\linewidth]{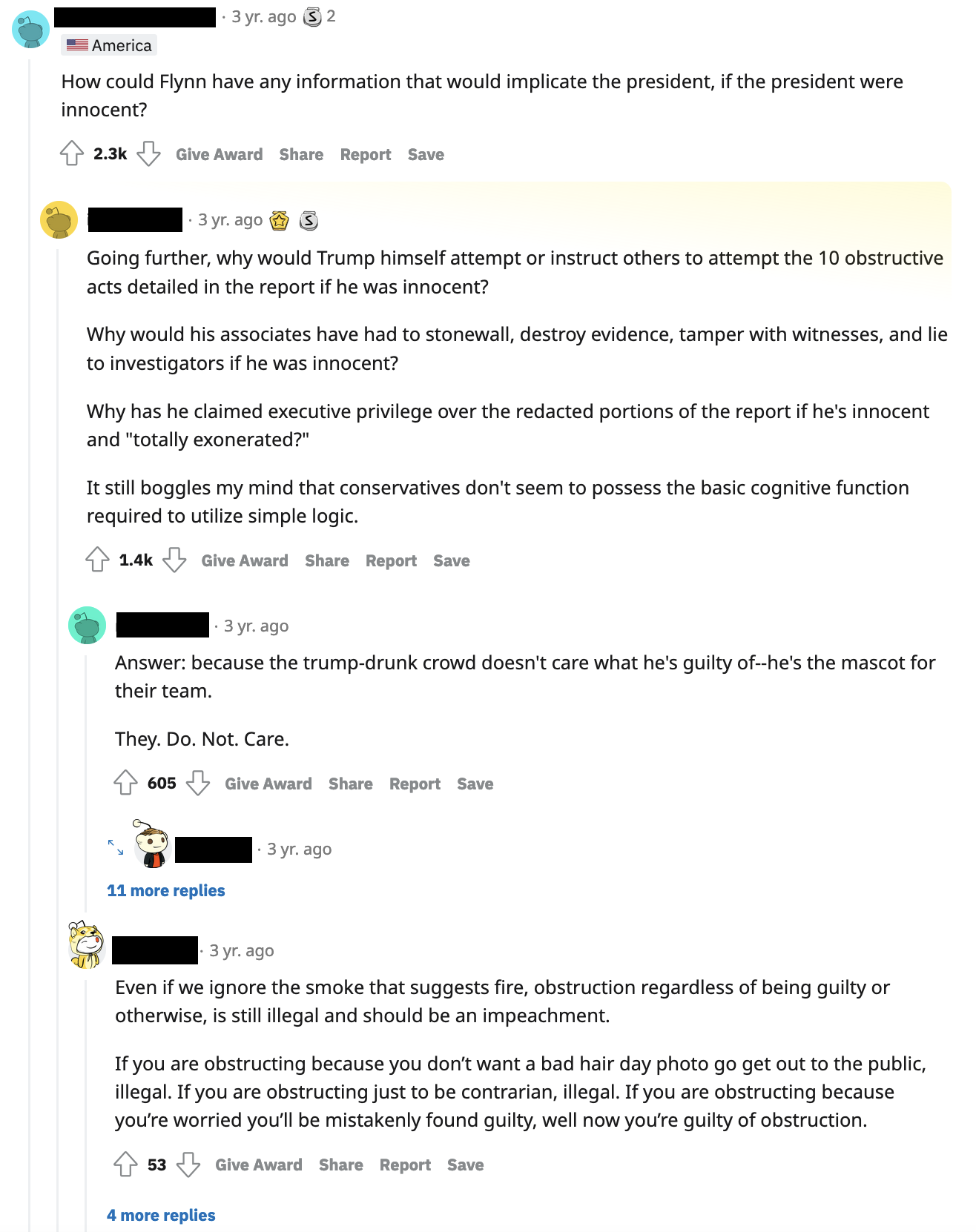}
    \caption{Example of a Reddit Discussion}
    \label{fig:reddit}
\end{figure}

The need for a deeper analysis of discussions with critical consideration of the relevant context becomes apparent when examining the excerpt from Reddit provided in Figure \ref{fig:reddit}. Focusing on the first interaction is insufficient in properly detecting the tendency towards hate, as these initial comments are actually quite benign. Indeed, what is perhaps most troubling with this excerpt is the kind of snowball effect of anti-social sentiment, as the discussion progresses.

In this work, we propose a novel approach to prevent hate speech on social platforms which addresses these challenges. First, it is important to properly encode the complex lexicon unique to online discussions \cite{belli2020privacyaware}. To do this, we utilize a fine-tuned BERT language model that was trained on hateful text from various social media platforms \cite{mathew2020hatexplain, devlin2019bert}. This allows our system to utilize contextual language embeddings of each node in the discussion graph as initial feature representations. Next, to capture the structure of conversations, we propose adaptations to Graphormer \cite{ying2021transformers}, a state-of-the-art graph transformer network. We adapt Graphormer away from its original purpose of detecting global qualities of molecules, a large difference from our intended use case. Instead, we 
predict whether specific nodes in our discussion graph will lead to discussions that encourage and contain hateful behaviour. By adapting this model, we benefit from a self-attention aggregation operation over the entire graph, allowing the model to learn contextual relationships between relevant comments in the discussion. Our end-to-end solution utilizes the embeddings generated by the language model as the representation of each vertex in our discussion graph, which are then aggregated and transformed by our Graphormer model to make final node-level predictions.

To train such a system, we assemble a dataset of 333,487 discussion graphs from different Reddit communities. Each node in a discussion tree is then algorithmically labeled with an ordinal value (0-4) according to how prevalent the hate speech is within the discussion that follows under that node. By utilizing a ranked representation of hate, we recognize that encouraged hateful behaviour can be more damaging than discouraged hateful behaviour, and that not all hateful behaviour is equivalent. In addition, users can leverage these scores to specify which level of hateful content they are comfortable viewing and can be used to help social platforms prioritize moderation for the most extreme content. We then investigate the impact of modeling specific communal influence by measuring performance of our system against variants that are trained on only one Reddit community. 

We conclude this work with a reflection on how our proactive approach to predicting hate speech can be used to improve the mental health of users online. This final discussion will shed more light on the significant value we derive from considering trees of discussion in social media, and not simply interpreting whether a single comment is conveying hate speech. Our codebase is available at \texttt{github.com/liamhebert/CommunityHateFormer}.

\section{Background and Related Work}
\label{related}

\subsection{Deep Language Models}
\label{sec:deep-language-models}
Our approach draws on recent advancements in deep language models and graph neural networks. To encode the textual content of a comment, we make use of the Transformer language model architecture proposed by Vaswani et al.~\cite{vaswani2017attention}. This architecture has resulted in many performance gains in various natural language tasks \cite{vaswani2017attention, devlin2019bert}.

Core to the transformer is the scaled dot product self-attention mechanism. Let $H^k = [h_0, h_1, ..., h_n] \in R^{n \times d}$ be the set of hidden embeddings of each word $h_i$ of size $d$ at layer $k$. Given three learnable weight matrices, $W_Q \in R^{d \times d}$, $W_K \in R^ {d \times d}$ and $W_V \in R^ {d \times d}$, the self-attention operation computes  
\begin{align*} 
    Q = H^{(k)}W_Q, ~K = H^{(k)}W_K, ~V = H^{(k)}W_V, \\
    Attn(H^{(k + 1)}) = softmax(\frac{QK^\top}{\sqrt{d}})V
\end{align*}
where $Attn(H^k)$ represents an attention-weighed representation of the input. This representation is then processed by a feedforward neural network to create $H^{(k + 1)}$ \cite{vaswani2017attention, ying2021transformers}. The use of a self-attention mechanism allows the model to encode the contextual relationship between words in a sentence \cite{vaswani2017attention}. BERT, proposed by Devlin et al., extends this architecture by adding a special token \texttt{[CLS]} to the initial input and using the final embedding $h_{[CLS]}$ to encode the entire sentence \cite{devlin2019bert}.

Transformer models can be fined-tuned to perform a variety of tasks, including hate speech detection on social media. To utilize this architecture towards hate speech detection, Mathew et al. curated a corpus of comments from social platforms Twitter and Gab, labeled as containing either Normal, Hateful or Offensive content \cite{mathew2020hatexplain}. The authors then fine-tuned a BERT model on this corpus to classify comments into these three categories. Their final system, BERT-HateXplain, achieved $\approx 70\%$ classification accuracy according to these three classes \cite{mathew2020hatexplain}. \edit{Our approach utilizes the $h_{[CLS]}$ embeddings created from a pre-trained BERT-HateXplain model to initially encode each comment in a discussion graph.} 

\subsection{Graph Neural Networks}
\label{sec:gnn}
\begin{figure}
    \centering
    \includegraphics[width=\linewidth]{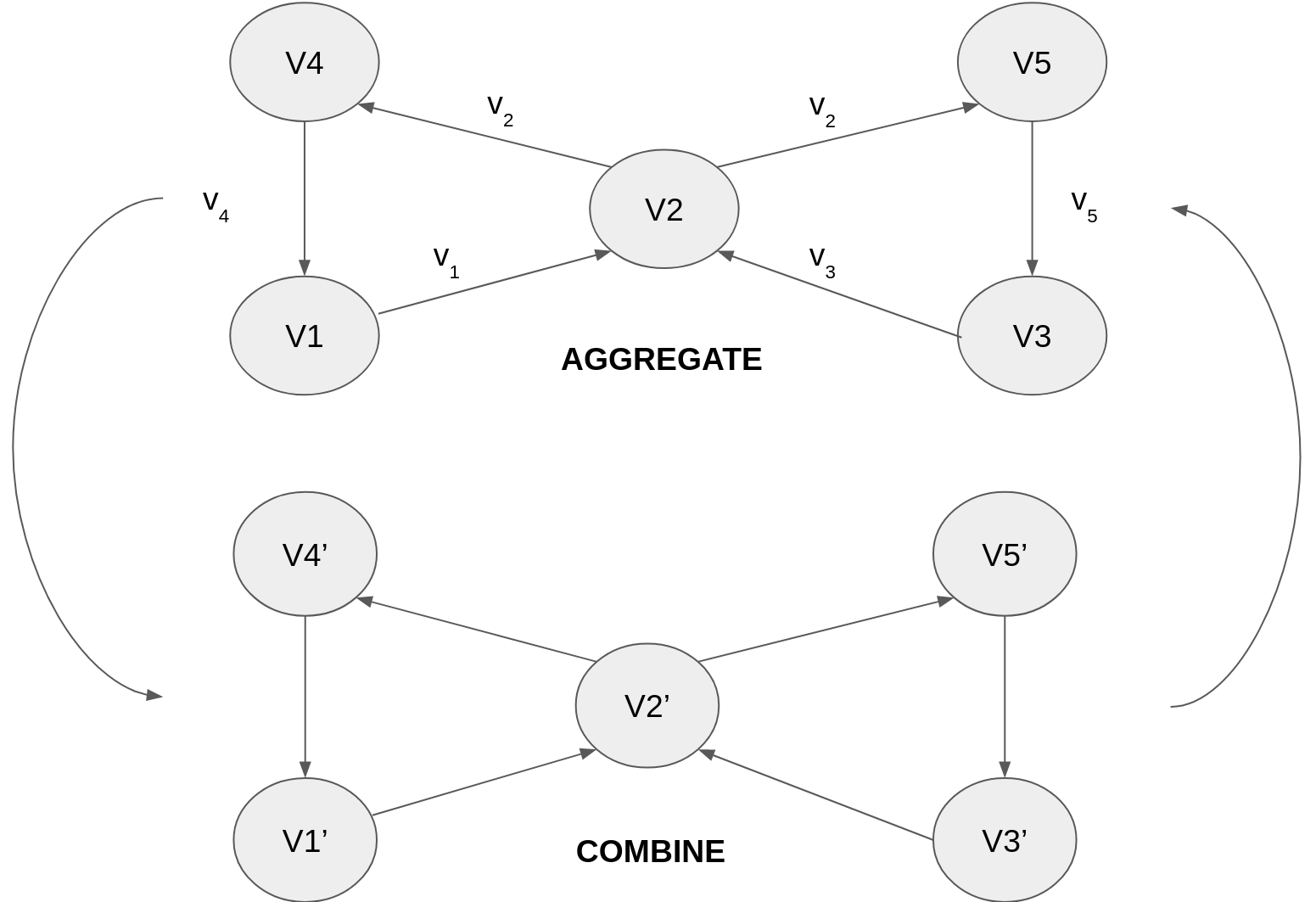}
    \caption{Aggregate-Combine Operations in a Graph Neural Network}
    \label{fig:aggregate_combine}
\end{figure}
Given a Graph $G = (V, E)$ with vertices $V$ connected by edges $E$, Graph Neural Networks (GNNs) compute representations of the vertices based on their feature embeddings $v$. Each layer of a GNN consists of two steps: AGGREGATE and COMBINE (Figure \ref{fig:aggregate_combine}). The AGGREGATE operator combines the feature embeddings $v_k$ of all neighbouring vertices of a given vertex $Vi$ into an aggregated embedding $a_i$. Common aggregation operators include taking the mean, max and sum of the feature embeddings of neighbouring vertices \cite{hamilton2018inductive, velickovic2018graph, kipf2017semisupervised, xu2019powerful}.

Following the AGGREGATE operator, the COMBINE operator creates a new representation of $Vi$ by transforming $v_i$ with respect to $a_i$. This new representation can then be used in subsequent layers for further aggregation. In practice, COMBINE operations are typically implemented as feed-forward neural networks,  which transform the concatenation of $a_i$ and $v_i$ \cite{ying2021transformers}. Key to the expressiveness of GNNs is that both the COMBINE and AGGREGATE steps are structure-independent. It is trivial to add vertices and edges to a GNN input and compute new inferences by simply expanding the number of vertices in the AGGREGATE operator. This aspect allows GNNs to be well suited for inputs that can take multiple shapes, such as branching social media conversations. 

\begin{figure}
    \centering
    \includegraphics[width=\linewidth]{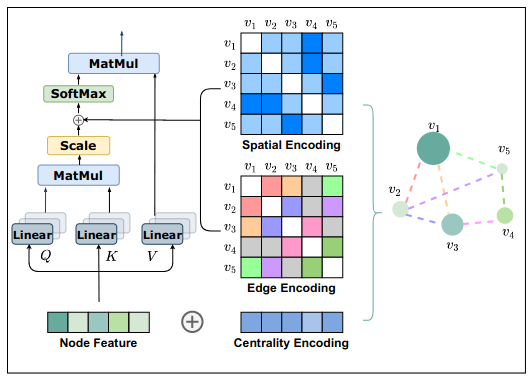}
    \caption{Graphormer Architecture \cite{ying2021transformers}}
    \label{fig:graphormer}
\end{figure}

Motivated by the performance increases achieved by Transformer models, Ying et al. adapted the self-attention mechanism to GNNs \cite{ying2021transformers} for use in molecular modeling, which they name Graphormer. A key feature of this architecture is the ability to use the self-attention mechanism to encode relationships between any nodes in the entire graph during each model layer, regardless of position and structure (Figure \ref{fig:graphormer}). To encode graph structure, Ying et al. propose four proxy variables. Centrality Encoding, representing the in-degree and out-degree of each node, is given as 
$$
H_i^{(0)} = x_i + z_{deg^-(v_i)}^- + z_{deg^+(v_i)}^+
$$
where $H_i^{(0)}$ is the initial representation of node $v_i$ with features $x_i$ and where $z_{deg^-(v_i)}^-$ and $z_{deg^-(v_i)}^+$ are learnable embeddings corresponding to the in-degree $deg^-(v_i)$ and out-degree $deg^+(v_i)$ of that node. These features are only added to the initial representations of each node. Next, Spatial Encoding is introduced to encode the structure of the graph according to inter-node distance. This encoding is added during the self attention mechanism, where the encoding of node $v_i$ with respect to node $v_j$ is given as 
$$
A_{ij} = \frac{(H_iW_Q)(H_jW_K)^T}{\sqrt{d}} + b_{\phi(v_i, v_j)}
$$
where $H_i$ and $H_j$ are the embeddings of nodes $v_i$ and $v_j$, $W_Q$ and $W_K$ are learnable weight matrices and $b$ is a learnable embedding representing the shortest distance between $v_i$ and $v_j$ in the graph ($\phi(v_i, v_j)$). The last two features are Edge Encoding, which encodes features corresponding to edge features that connect nodes, and \texttt{[CLS]} nodes, to represent the entire structure. Following the self-attention mechanism, the next layer representation $H^{(k + 1)}_i$ of node $i$ is given as 
$$
H^{(k + 1)}_i = softmax(A_i) \times H^{(k)}W_v 
$$
where $W_v$ is a learnable weight matrix and $softmax(A_i)$ is the softmax operation over all $A_i$ embeddings. These $A_i$ embeddings are created for every node pair in the graph regardless of position, allowing node embeddings to be created with respect to the entire graph structure.

\subsection{Prior Work}
The urgency of detecting hate speech online has led to a large research interest in natural language processing tools. However, \textit{preventative} hate speech detection is still a difficult problem to solve. Prior work on this problem by Brassard-Gourdeau et al. proposed utilizing features from the first two messages of a linear conversation, in addition to features in the current message, as useful predictors for future hate speech \cite{Brassard2021Using}. Parallel work by Huang et al. 
proposed methods to pre-filter comments before being published to a wider audience \cite{huang-etal-2018-cyberbullying}. However, both of these systems operate on a limited context, narrowed to the first few messages of a discussion or on the next comment to be made. In addition, many of these systems are trained to generalize to all kinds of social discourse, which may hinder the detection of nuanced culture-specific hate speech. A final concern is that the systems are often designed to handle linear discussions, whereas online social media typically have branching discussions instead. 

GNNs have recently been proposed to capture the importance of context to predict hateful discourse in complex discussion graphs. Parmentier et al. investigated using GNNs to detect traits of individual comments by aggregating features of their immediate neighbours \cite{parmentier2021}. This included traits such as predicting the ``upvote'' and ``downvote'' score of a comment and whether the comment contained hateful content \cite{parmentier2019learning, parmentier2021}. However, while predicting the behaviour stemming from a comment, Parmentier et al. only examined immediate node neighbours.
This was a constraint originating from the methods and computational resources available at the time. 

We differ from this prior work in two ways. First, instead of predicting qualities of individual nodes, we predict whether specific nodes will lead to future hateful discourse by consuming the entire discussion graph during inference. To do this, we treat predictions of each node in the graph as the prediction of the future subtree rooted at that node. This allows our system to be utilized towards preventative detection of hate speech, rather then analyzing comments after they have been published. In addition, the self-attention mechanism allows our model to only pay attention to contextually relevant nodes when computing new representations, avoiding noise. 

Second, we differ from this prior work by incorporating fine-tuned contextual deep language models rather than using context-free GloVE language modeling \cite{pennington2014glove}. By utilizing deep language models, we create embeddings that are fine-tuned on the lexicon that is present on online social platforms, rather than relying on embeddings generated from formal sources, such as Wikipedia. Additionally, self-attention allows the language model to contextualize to the entire input, as opposed to GloVE which relies on word co-occurrence ratios of the training corpus \cite{pennington2014glove}. Notably, the usage of self-attention mechanisms  resulted in much better performance in context-dependent tasks, such as sentiment analysis \cite{vaswani2017attention}. In summary, the novelty of our approach lies on combining deep language models with recent innovations in GNNs that can efficiently capture longer range relationships. 

\section{Methodology}
\label{methodology}

\subsection{RedditHate Dataset}
\label{sec:reddit-hate}
To train our model, we collected Reddit posts created between 2018 and 2019 using the Pushshift  API\footnote{https://github.com/pushshift/api}, focusing on the \texttt{The\_Donald}, \texttt{IAmA}, \texttt{AmITheAsshole} and \texttt{Politics} communities. This resulted in a dataset of 333,487 posts and 6,531,455 comments. These communities were selected due to their frequent lively discussion around various topics and known differences in community norms \cite{parmentier2019learning, Waller2021}. 

Due to the size of the dataset, it is not feasible to label every subtree by hand. Instead, we rely on several proxy features. First, we make use of the scores (upvotes and downvotes) that users assign to comments on Reddit. We utilize an aggregated tally of these scores to model how inflammatory a given comment was to a community. \edit{We also make use of the hatefulness prediction of comment text from BERT Hate-Explain \cite{mathew2020hatexplain} (retaining the definition of hate by the authors) and the graph structure as further proxy variables. For each vertex $v_i$ in the graph, we compute the hate label of a comment $L_{v_i}$ by summing over three symbolic terms that estimate how hateful the conversation is stemming from that node: Context $C_{v_i}$, Reaction $R_{v_i}$ and Influence $I_{v_i}$. These terms are computed as:} 
\begin{align} 
    C_{v_i} &= 0.25 \times hate_{p_i} \times score_{p_i} \\
    R_{v_i} &= 0.25 \times hate_{v_i} \times score_{v_i} \\
    I_{v_i} &= \sum_{c\in S_v} 0.25 \times L_c \\
    L_{v_i} &= C_{v_i} + R_{v_i} + I_{v_i}
\end{align}
where $hate_v$ is the HateXplain predicted score of vertex $v$ scaled to $(-0.7 - 1.5)$, $score_v$ is the upvote/downvote score, $p$ is the parent node of $v$ and $S_v$ is the set of child nodes under $v$. The Context term $C_{v_i}$ aims to capture the context in which a comment was made, focusing on the hatefulness of the parent comment. The Reaction $R_{v_i}$ term then captures the hatefulness of the instigating comment that led to the conversation that followed. Lastly, the Influence term $I_{v_i}$ recursively sums the discounted hatefulness of the conversation that followed as a result of the current node. Each label is created in a bottom-up approach starting from the leaves of the discussion tree. \edit{The factor 0.25 in each formula is selected to equally weigh the contribution of each component.} As part of our design process, we conducted a sensitivity analysis of these weights, detailed in Section \ref{sec:sensitive}, and found no significant differences in performance when perturbed.  

In each term comprising our hate label $L_{v_i}$ (Equations 1 through 3), the multiplication ensures that hateful posts that are encouraged with a positive score have a similar score to non-hateful posts that have a negative score. Additionally, the weights attached to each term allow lively hateful discussions with many subtrees to decisively influence the label of the current comment. We then standardize each label into five ordinal classes, ([$<0$], [0 - 5], [6 - 20], [21 - 500], [$500<$]), corresponding to increasing levels of hatefulness. \edit{The intuition behind the use of ordinal values is to enable social media moderators to prioritize the most hateful content and deploy different policies for different criteria. The range of values chosen for each of these classes was selected as a result of inspection of the overall distribution of hate labels and the comments contained within them. Our classification of hate can then be seen as a regression problem.} The final label distribution can be seen in Table \ref{tab:label_distribution}. With each discussion tree labeled, we then convert each graph into Pytorch Geometric (PyG) Data Graphs objects for inference \cite{fey2019fast}. 

This approach of using algorithmic labeling follows prior work for labeling hate speech on social platforms, such as \cite{parmentier2019learning} and \cite{Brassard2021Using}, among other works. In each of these systems, algorithmic methods such as HateXplain replace human reviewers 
to label a sufficiently large dataset. 

Next, to analyze the ability of our model to proactively predict hate speech, we take a subset of each labeled discussion tree as the model's input. We do this by taking a horizontal slice of the discussion tree such that the maximum depth of the tree is four comments from the root. We then remove all comments that did not lead to a sub-discussion of at least two comments. As a result, the leaf nodes of our trimmed discussion tree are parent nodes of at least two other comments, requiring the model to predict the hatefulness of an unseen future discussion. Our trimming strategy also focuses our model's input on the most relevant parts of the discussion \cite{parmentier2021}. Table~\ref{tab:sample_size} shows the number of nodes before and after filtering for the four communities we selected. 

\begin{table}[t]
    \centering
        \caption{Community Sample Size for the RedditHate Dataset}
    \begin{tabular}{c|c|c}
        Subreddit & Nodes & Nodes after Filtering \\
        \hline
        The\_Donald & 6 531 455 & 3 005 612\\
        Politics & 6 127 723 & 1 457 321\\
        AmItheAsshole & 4 135 270 & 1 381 200\\
        IAmA & 183 284 & 30 415\\
    \end{tabular}
    \vspace{0.2cm}

    \label{tab:sample_size}
\end{table}

\begin{table}[t]
    \centering
        \caption{RedditHate Label Distribution}
    \begin{tabular}{c|c|c|c|c|c}
        Subreddit / Label & 0 & 1 & 2 & 3 & 4 \\
        \hline\hline
        AmItheAsshole & 1 103 308 & 183 108 & 39 815 & 45 371 & 9598 \\
        IAmA & 27 890 & 2 240 & 216 & 68 & 1 \\
        The\_Donald & 2 310 601 & 441 746 & 133 534 & 117 091 & 2640 \\ 
        Politics & 1 087 286 & 264 570 & 54 311 & 45 563 & 5 591 \\
        \hline
        Total & 4 529 085 & 891 664 & 227 876 & 208 093 & 17 830
    \end{tabular}

    \label{tab:label_distribution}
\end{table}

\subsection{Model Architecture}
Our approach modifies the Graphormer model by Ying et al.~\cite{ying2021transformers} with the HateXplain model by Mathew et al.~\cite{mathew2020hatexplain}. Notably, the original Graphormer architecture was proposed and geared towards molecular modeling and graph-level predictions (Section \ref{sec:gnn}). However, with respect to the RedditHate Dataset (Section \ref{sec:reddit-hate}), our task is instead to have node-level predictions regarding the future direction of discussions. Despite this, our focus on capturing the entire discussion context makes the Graphormer architecture appealing.

To adapt the Graphormer architecture towards our task, we first omit the usage of \texttt{[CLS]} nodes and Edge Encoding since we are not concerned with graph level predictions or edge features. Second, we adapt the model output to predict ordinal labels for each node in the graph. These labels correspond to the predicted hate score defined in Section \ref{sec:reddit-hate}. Finally, to encode each comment in the graph, we utilize a pre-trained BERT model trained on the HateXplain dataset (Section  \ref{sec:deep-language-models}). We use the \texttt{[CLS]} embedding of each comment as the feature representation in the graph. As a result, each comment in the discussion graph is encoded as a 769-dimension embedding. The usage of BERT-HateXplain allows each contextual comment embedding to be learned from common social media discourse, rather than formal and clean text. We also concatenate the upvote/downvote score to the input vector. \edit{In our experiments, we utilize the Graphormer-BASE variant which consists of ten layers of Graphormer self-attention, resulting in 144 million total parameters (36 million excluding BERT-HateXplain)}.

\section{Results}
In our experimental evaluation, we start by investigating whether global discussion context allows for more accurate detection of future hateful discussions (results in Section~\ref{sec:exp_context}). This is motivated by contrasting against other attempts at detecting hate speech, which often rely on just a single comment or the closest neighbours of a comment \cite{parmentier2021, Brassard2021Using, huang-etal-2018-cyberbullying}. 
Second, under the lens of modeling diverse discussion semantics, we investigate whether models that are trained on specific communities outperform models that are trained on all communities (results in Section~\ref{sec:exp_community}). Expanding on these questions, in Section~\ref{sec:exp_qualitative} we present a qualitative comparison of our approach against existing approaches, giving examples of discussions that were predicted more effectively by our method, compared to others. We then conclude with a sensitivity analysis of our hate labelling function in Section~\ref{sec:sensitive}. 

We use the standard Graphormer variant, which consists of ten transformer layers with a hidden dimension size of 769, as described in the original work. Additionally, we utilized the AdamW optimizer with a peak learning rate of 2e-4, which linearly decayed after a warmup period \cite{loshchilov2019decoupled}. \edit{We applied an L2 regression loss} for training and as an evaluation metric, which assigns heavier penalties towards incorrect values. This is important due to the imbalance in our dataset towards smaller hate label values. Due to compute constraints, a batch size of 16 was used over ten epochs and only the Graphormer model was fine-tuned. To evaluate the performance of the model, we masked 10\% of comments for use in validation and 20\% for use in testing, following prior work by Chen et al. \cite{chen2018fastgcn}. \edit{We utilize the validation set for hyper-parameter tuning in our experiments}. Each model was trained for $\approx 5$ hours with 3 RTX 2080 Ti GPUs. 
 
 \subsection{Impact of Deeper Context}
\label{sec:exp_context}

\begin{table}[t]
    \centering
        \caption{Model Performance (L2 Loss) of our approach, labelled Graphormer, and GAT}
    \begin{tabular}{c|c||c|c|c|c|c}  
        Method / Label & All & 0 & 1 & 2 & 3 & 4\\
        \hline
        Graphormer & 1.066 & 0.556 & 0.946 & 1.275 & 2.213 & 3.817 \\
        GAT & 1.485 & 0.473 & 1.228 & 1.872 & 2.661 & 5.883  \\
    \end{tabular}
    \label{tab:gat}
\end{table}

We begin by investigating the impact of including the entire discussion graph during inference. To measure this, we selected the Graph Attention Network architecture (GAT) \cite{velickovic2018graph} as a baseline, as implemented in Pytorch Geometric \cite{fey2019fast}. This is inspired by previous efforts to utilize graph models to detect hate speech on social networks, in particular \cite{parmentier2021}, which utilized this architecture. It is important to note that our use case differs, where we instead seek to detect comments that will \textit{lead} to hateful discussion. Similar to the Graphormer architecture, the GAT architecture utilizes an attention mechanism to aggregate neighbouring node embeddings. However, an important difference with this architecture is that the aggregation operation is constrained to the direct neighbouring comments in each layer. This is contrasted by our proposed approach, which performs self-attention aggregation over the entire graph at each layer. 

We compare our approach against a two-layer GAT model, following the recommended configuration by the original authors and prior work \cite{parmentier2021}. This implies two iterations of the Aggregation-Combine operation. Table \ref{tab:gat} shows the results: our approach (Graphormer) outperforms GAT by an average L2 loss of 0.419. Our results illustrate that both models perform well at classifying comments with a low level of future hateful discussion (0-1). This can be attributed to the prevalence of these comments in our dataset. However, notable differences in model performance can be seen when detecting comments which led to more prevalent hatespeech (2-4), where our model outperforms GAT by an average L2 loss of 2.066. 

We hypothesize that the gains made by Graphormer are due to its ability to contextualize better to the discourse of larger discussion graphs, which are often good indicators of future lively discussions. An example of this would be an instigating comment in a debate, where a better understanding of the entire debate could serve as a better predictor of what could be considered hateful and instigating rebuttal. That being said, it is still important to note that the performance at the higher echelon is not optimal, having an average prediction inaccuracy of $\approx 2$ ordinal levels; however, Graphormer still presents an improvement over GAT. 

\subsection{Impact of Community Specific Models}
\label{sec:exp_community}
\label{results}
\begin{table}
    \centering
        \caption{Model L2Loss Performance on Different Communities}
    \begin{tabular}{c|c|c}
        Target Communities & Train & Test \\
        \hline\hline
        All & 1.060 & 1.066 \\
        The\_Donald & 0.4913 & 0.4911 \\
        Politics & 0.4042 & 0.4041 \\ 
        IAmA & 0.3687 & 0.3444 \\ 
        AmITheAsshole & 0.4388 & 0.4397 \\
    \end{tabular}
    \label{tab:model_performance}
\end{table}
A key question in this work was whether hate could be better predicted by modeling the behaviours and cultural norms of specific communities. To evaluate this, we trained variations of the model on specific communities present in our dataset. We compare these results against a general purpose variant that was trained on data from all four communities (Table \ref{tab:model_performance}). We see that general models trained on all subreddits perform significantly worse than community-specific models. When comparing L2 loss, each community-specific model outperforms general models by a factor of two, for an average loss of $\approx 0.4$. The model for the \texttt{IAmA} subreddit benefited the most from having a community-specific focus, with an average L2 loss of 0.3687, followed by the \texttt{politics} subreddit.

We hypothesize that this performance boost came from two advantages. First, each community we selected is associated with vastly different topics, with \texttt{The\_Donald} created to discuss alt-right politics and \texttt{AmITheAsshole} discussing personal advice. As such, it is likely that comments that would be ignored in \texttt{The\_Donald} could be inflammatory in other communities. A similar trend can be noticed in \texttt{politics}, which is politically left leaning and would be more sensitive to hateful political discourse \cite{Waller2021}. By modeling specific communities, we hypothesize that our model captured these different cultural norms.

Second, some communities discuss in a formal and structured manner, such as question-answer format. \texttt{IAmA} is an example of this, a community that focuses on interviewing celebrities by asking public questions, contrasting other communities which have free-flowing discussion. This difference was highlighted in our results, where we found that \texttt{IAmA} benefited the most from a community-specific approach. By modeling the question and answer format and by sampling the entire interview, we hypothesize that our model was able to better understand the topics discussed and the unique conversational flow. 


\subsection{Qualitative Analysis}
\label{sec:exp_qualitative}

\begin{table*}
    \centering
        \caption{Example conversation requiring long range forecasting from community cues (/r/politics) \\ * indicates comments outside of initial input context and predicted in subsequent iterations}
    \begin{tabular}{c|p{9cm}|c|c|c|c}
        Depth & Text & Label & Graphormer & GAT & Bert-HateXplain  \\
        \hline
        \hline
        0 & How could Flynn have any information that would implicate the president, if the president were innocent? & 3 & \textbf{4} & 1 & 0 \\
        \hline
        1 & Because, according to the GOP Hive Mind, the information is a combination of completely fabricated and not illegal. Trump's narrative has always been "Witch Hunt!" i.e., that there was no basis for the investigation, so any information is obviously fake. [...] It still boggles my mind that conservatives don't seem to possess the basic cognitive function required to utilize simple logic. & 3 & 3 & 2 & 1 \\
        \hline
        1 & Going further, why would Trump himself attempt or instruct others to attempt the 10 obstructive acts detailed in the report if he was innocent? Why would his associates have had to stonewall, destroy evidence, tamper with witnesses, and lie to investigators if he was innocent? & 2 & 3 & 0 & 1 \\
        \hline
          & [...] & & \\
        \hline
        6 & If Republicans continue the trend, their next candidate will only be grunting and throwing feces. & 4 & \textbf{4}* & 4* & 4*\\
        \hline
        7 & I think we can go dumber.. & 4 & \textbf{3}* & \textbf{1}* & 2*
    \end{tabular}

    \label{tab:predictive}
\end{table*}
To better capture the behaviours of our  approach, we conducted a qualitative analysis against GAT and comment-text only Bert-HateXplain. Since the Bert-HateXplain values are originally between 0-1, we map them to increasing ordinal values of [0-0.2], [0.2-0.4], [0.4-0.6], [0.6-0.8] and [0.8-1.0], for a matching 0-4 scale. We focused on investigating conversations where predictions between all three models differed. An example of such a conversation, found in the \texttt{politics} subreddit, is shown in Table $\ref{tab:predictive}$. The subject of this conversation is regarding U.S. politics and then-recent news regarding Michael Flynn, former security advisor to President Donald Trump. On the \texttt{politics} subreddit, prior work has discovered a strong left leaning bias, resulting in heated discussion when talking about topics concerning the political right \cite{Waller2021}. In this discussion, the conversation starts with negative but still relatively tame comments. However, as the conversation deepens, it devolves into hate speech against the Republican party ("If Republicans continue the trend, their next candidate will only be grunting and throwing feces"). 

Investigating the predicted values of all three methods, we see that Graphormer more accurately predict the direction the discussion will be heading, even stemming from the first comment. Our approach predicted an ordinal score of 4 compared to a label of 3. This is compared to predictions made from GAT and HateXplain, which fail to capture the longer dependency (1 and 0 respectively). We also see a recurring trend with single comment Bert-HateXplain, which seems to downplay the intensity of the hateful content early on compared to the ground truth labels. \edit{It is important to note that the input discussion is cut off after a depth of four comments (the limit used in \cite{parmentier2021}), and therefore predictions must be made from that limited context.  }

\subsection{Sensitivity of RedditHate Labeling Function}
\label{sec:sensitive}

\begin{table}
    \centering
        \caption{Model Performance on Different Labeling Weights}
    \begin{tabular}{c|c|c}
        Labeling Weight Variant & Train L2Loss & Test L2Loss \\
        \hline\hline
        Equal & 1.060 & 1.066 \\
        Influence-Weighted & 1.180 & 1.200 \\
        Reaction-Weighted & 1.072 & 1.073 \\ 
        Context-Weighted & 0.985 & 0.994 \\ 
    \end{tabular}

    \label{tab:model_sensitivity}
\end{table}
We now investigate the sensitivity of the weights defined in the HateReddit labeling function, which combines the value of the Context-term (original node focused), Reaction-term (parent node focused) and Influence-term (aggregated child nodes focused) (Section \ref{sec:reddit-hate}). This evaluates the likelihood of our results to be dependent on the labeling weights chosen. To do this, we created four variants of the dataset based on different weighting configurations: Equal, which consists of weighting each term equally, as well as Influence-Weighted, Reaction-Weighted and Context-Weighted, which doubles the weight of each respective term in the labeling function (weighted by 0.5 instead of 0.25). We conducted each experiment on the entirety of the HateReddit dataset. 

Table \ref{tab:model_sensitivity} shows the results. Each variant performs  similarly, with a Loss range of (0.994 - 1.200). As a result, we conclude that the labeling weights are not sensitive to specific values. We see that the Context-Weighted variant of the dataset produced the best performance with a Test L2 loss of 0.994, whereas the worst performing variant was the Influence-Weighted variant, with a Test L2 loss of 1.200. It is important to note that the Influence term of each node is calculated from the entire discussion tree, despite the input constrained to a depth of four with each remaining comment being the parent of at least two others, potentially hidden, comments. As a result, the Influence-weighted variant represents a more difficult task compared to the other variants. 

\section{Future Work}
\label{futurework}
There are several potential avenues for future work. First, recall that the RedditHate dataset is founded on algorithmic deduction that approximates ground truth. Ideally, the RedditHate dataset would be annotated by human evaluators or more advanced labeling strategies.
Potential future work could also explore creating an expanded and balanced dataset of hateful and non-hateful discussions. This could be done by considering a longer time period and then carefully selecting discussions to fit a desired distribution. In addition, the collection of communities discussed could be further expanded to consider more diverse and topic orientated communities. 

Second, we hypothesize that it would be beneficial to train the model end-to-end by fine-tuning the BERT-HateXplain and Graphormer models together. In this work, we circumvented the need to train the BERT model by using a publicly available pre-trained variant that was trained on Twitter and Gab data towards detecting hateful comments. By fine-tuning a model directly on comments from specific communities and towards our target task, we can create embeddings that directly reflect the behaviours of the target community. We believe that this could allow for better subreddit-specific performance.  

Third, future work could investigate the interpretability of the graph attention weights during inference, following past work in NLP \cite{vashishth2019attention}. This could enable an exploration of the kinds of comments that enable hateful discussions in specific communities, and the communal influence that enables them.

Fourth, further analysis could be performed to compare community-specific models. By analyzing the performance of one community model on posts from a different community, it would be possible to derive insights on cultural similarities. That is, models that perform well on other communities likely share similar discussion norms. An example of such a comparison would be evaluating the performance of the \texttt{politics} model on \texttt{the\_donald} subreddit, communities that are known to reflect contrasting political biases. This direction could also expand to applying our model on other
social platforms, such as Twitter, in order to learn whether the performance of the approach is equally strong in
all environments. We will need to decide how best to model discussions and communities
within a platform such as Twitter, in order to generate our results.

Lastly, it would be beneficial to conduct additional experiments in order to draw out the real strength of our predictive approach to hate detection, in comparison with simpler baselines which rely on simply tagging existing comments. We would code a competitor which assumes that the initial post sets the tone for the overall conversation of each discussion (i.e., all comments that follow this post are labelled with the same level of hate detection). We would then examine the performance of this method for detecting hate speech in Reddit, both across all subreddits as well as in the specific communities listed in Table IV, where the performance of our own approach in terms of L2 loss has already been measured. We would expect to see, for the baseline, significant degradation on each subreddit. This would then confirm that the initial post is often not indicative to predict the hateful labels of the conversation and thus that our approach of incorporating attention to predict hate patterns has significant value. For future work, we can also examine other competitors that fail to incorporate context, perhaps in other social media environments, to continue to calibrate the relative benefit of our solution.

\section{Conclusion}
\label{conclusion}
Social media have increasingly become the source of many damaging mental health effects \cite{2020jungling}. In this work, we proposed a system based on state-of-the-art graph transformer models and deep language models to prevent the spread of harmful discourse that has perpetuated these damaging effects. To evaluate our system, we created a dataset that contains a collection of Reddit posts from various communities. We hope that our system can be used by social media moderators to curb the spread of harmful discourse by motivating the usage of community specific models and the importance of capturing discourse context.

The scope of this work differs in several ways from previous work. First, we studied the influence of communities regarding what content is shared and how that content is reacted to. We accomplished this by including the entire discussion graph during inference and creating specific models for individual communities. This scope is different from traditional hate speech systems that only examine the text of a comment in isolation, without considering the discussion context or the community. As we have shown in our experimental results, including more context, in terms of the discussion and the community, can lead to better performance.

Second, the output of the model differed from previous work by predicting ordinal values from 0 to 4, denoting how intense and encouraged the hateful discussion will be following an initial comment. There are two benefits that come with this objective. First, due to the increasing scope of social platforms, it becomes important for moderators to prioritize which content to investigate first and which content should be immediately quarantined. By \edit{including a range of intensity values, we also support users of these platforms to choose an intensity level that they are comfortable with.} This would allow users with mental health challenges to self-regulate the content they want to see without restricting the free speech of users who are more comfortable seeing intense content.  

We also hope that the insights and tools created in this research can be used to drive future research into enhanced social media moderation under a proactive lens, especially as our lives and mental health become increasingly connected to social platforms. With a deeper exploration of an entire tree of discussion following a post and with a contextual attention mechanism that improves the efficiency of predicting what may be yet to come, our approach has the chance of detecting when a growing surge of negative expression may be unleashed. By adapting a more proactive and less reactive approach to hate speech, the hope is that fragile users may be better protected from the effects of anti-social behaviour. 

\section*{Acknowledgment}
\edit{The authors thank the Natural Sciences and Engineering Research Council of Canada, the Canada Research Chairs Program and the University of Waterloo Cheriton Scholarship for financial support. We are also grateful to the reviewers for their valued feedback on the paper.}
\bibliographystyle{IEEEtran}
\bibliography{references}

\end{document}